%% file: root.tex
\documentclass[letterpaper, 10 pt, conference]{ieeeconf}  % Comment this line out if you need a4paper

\IEEEoverridecommandlockouts                              % This command is only needed if 
\usepackage{amsmath,amsfonts}
\usepackage{amssymb}

\usepackage{algorithm}
\usepackage[noend]{algorithmic}
\usepackage{color}

\usepackage{array}
\usepackage{textcomp}
\usepackage{stfloats}

\usepackage{verbatim}
\usepackage{graphicx}

\usepackage{cite}

\usepackage[utf8]{inputenc} % allow utf-8 input
\usepackage[T1]{fontenc}    % use 8-bit T1 fonts
\usepackage{hyperref}       % hyperlinks
\usepackage{url}            % simple URL typesetting
\usepackage{booktabs}       % professional-quality tables
\usepackage{amsfonts}       % blackboard math symbols
\usepackage{nicefrac}       % compact symbols for 1/2, etc.
\usepackage{microtype}      % microtypography
\usepackage{lipsum}
\usepackage{fancyhdr}       % header
\usepackage{graphicx}       % graphics
\usepackage{wrapfig}
\usepackage{subfigure}
\usepackage{multirow}

\title{\LARGE \bf
Hybrid Consistency Policy: Decoupling Multi-Modal Diversity and Real-Time Efficiency in Robotic Manipulation
}

\author{
    Qianyou Zhao$^{1}$, Yuliang Shen$^{1}$, Xuanran Zhai$^{2}$, Ce Hao$^{2}$, Duidi Wu$^{1}$,  Jin Qi$^{1}$, Jie Hu$^{1\dagger}$, and Qiaojun Yu$^{3\dagger}$ 
    \thanks{$\dagger$ Corresponding Author: Qiaojun Yu, Email: yqjllxs@alumni.sjtu.edu.cn.}
    \thanks{$^1$ Qianyou Zhao, Yuliang Shen, Duidi Wu, Jin Qi, and Jie Hu are with Shanghai Jiao Tong University.}
    \thanks{$^2$ Xuanran Zhai and Ce Hao are with National University of Singapore.}
    \thanks{$^3$ Qiaojun Yu is with Shanghai AI Lab.}
}

\begin{document}

\maketitle
\thispagestyle{empty}
\pagestyle{empty}

%%%%%%%%%%%%%%%%%%%%%%%%%%%%%%%%%%%%%%%%%%%%%%%%%%%%%%%%%%%%%%%%%%%%%%%%%%%%%%%%

\input{sections/0_Abstract}
\input{sections/1_Introduction}

\input{sections/2_Relatedworks}
\input{sections/3_Preliminary}
\input{sections/4_Method}
\input{sections/5_Experiment}
\input{sections/6_Conclusion}

%%%%%%%%%%%%%%%%%%%%%%%%%%%%%%%%%%%%%%%%%%%%%%%%%%%%%%%%%%%%%%%%%%%%%%%%%%%%%%%%
\bibliographystyle{IEEEtran}
\bibliography{references}

\end{document}

%% file: sections/0_Abstract.tex
\begin{abstract}
% In visuomotor policy learning, diffusion-based SDE methods model multi-modal behaviors well but sample slowly, while ODE methods are much faster but often lose multi-modality. 
% In visuomotor policy learning, diffusion-based imitation learning has become widely adopted for its ability to capture diverse behaviors.  However, approaches built on Ordinary and Stochastic denoising processes struggle to jointly achieve fast sampling and strong multi-modality.
% To solve these challenges, we propose the Hybrid Consistency Policy (HCP). HCP runs a short stochastic SDE prefix up to an adaptive switch time 
% $t_s^\ast$, and then applies a one-step consistency jump to produce the final action. In both simulation and on a real robot, HCP with 25 SDE steps plus one jump approaches the 80-step DDPM teacher in accuracy and mode coverage while significantly reducing latency. These results show that multi-modality does not require slow inference, and a well-chosen skip time decouples mode retention from speed. It yields a practical accuracy–efficiency trade-off for robot policies. Project website: \href{https://sites.google.com/view/hybrid-cp}{https://sites.google.com/view/hybrid-cp}

In visuomotor policy learning, diffusion-based imitation learning has become widely adopted for its ability to capture diverse behaviors. However, approaches built on ordinary and stochastic denoising processes struggle to jointly achieve fast sampling and strong multi-modality.
To address these challenges, we propose the Hybrid Consistency Policy (HCP). HCP runs a short stochastic prefix up to an adaptive switch time, and then applies a one-step consistency jump to produce the final action.
To align this one-jump generation, HCP performs time-varying consistency distillation that combines a trajectory-consistency objective to keep neighboring predictions coherent and a denoising-matching objective to improve local fidelity.
In both simulation and on a real robot, HCP with 25 SDE steps plus one jump approaches the 80-step DDPM teacher in accuracy and mode coverage while significantly reducing latency. These results show that multi-modality does not require slow inference, and a switch time decouples mode retention from speed. It yields a practical accuracy–efficiency trade-off for robot policies. Project website: \href{https://sites.google.com/view/hybrid-cp}{https://sites.google.com/view/hybrid-cp}.
\end{abstract}

%% file: sections/1_Introduction.tex
\section{Introduction} \label{Sec: intro}

In recent years, diffusion-based methods have shown strong competitiveness in visuomotor control and sequential decision making \cite{wang2022diffusion,chen2023diffusion,chi2023diffusion,ze2024-3ddiffusion,reuss2023goaldiffusion,kim2024openvla}. In particular, Diffusion Policy (DP) \cite{chi2023diffusion} outperforms prior approaches on many manipulation benchmarks and real platforms \cite{zare2024survey}. It treats action sequences as the generative target and uses iterative denoising to achieve stable training and strong expressivity for multi-modal action distributions. DP based on Stochastic Differential Equation (SDE) \cite{ho2020denoising} requires tens to hundreds of denoising steps at inference, and reducing the steps degrades performance. The inference time limits DP and makes low-latency control on resource-constrained platforms difficult \cite{chi2023diffusion}.

Diffusion methods based on Ordinary Differential Equation (ODE) ~\cite{song2020ode} accelerate diffusion model inference \cite{wang2024one,consistency,sun2025flashback,zhang2025flowpolicy}. To relieve the slow sampling bottleneck, researchers adopt the ODE view of score based models \cite{song2020denoising,kim2023consistency,karras2022elucidating}. In theory, the reverse stochastic differential equation and its corresponding ODE share the same marginal distributions, and the ODE trajectory is deterministic. Along this way, methods based on ODE use a non-Markovian construction to reduce sampling steps and open a practical path for few-step sampling \cite{song2020denoising}. In robotics, Consistency Policy \cite{consistency}, ManiCM \cite{lu2024manicm}, and Flow Policy \cite{zhang2025flowpolicy} demonstrate fast policies with competitive success.

% Progressive distillation \cite{kim2023consistency} further compresses the number of steps by distilling a slow teacher into a fast student while largely preserving sample quality. Consistency Models (CM) \cite{pmlr-v202-song23a} are trained or distilled to enable one-step or few-step generation by enforcing agreement of denoising results that start from different locations along the same ODE trajectory. In robotics, Consistency Policy \cite{consistency} distills DP into a few-step student and accelerates inference by about an order of magnitude while retaining competitive success rates. Concurrent progress includes ManiCM \cite{lu2024manicm} for real-time manipulation with point clouds and FlowPolicy \cite{zhang2025flowpolicy} based on consistency flow matching. These methods achieve fast policy generation under different perception and action parameterizations and further support the suitability of the ODE paradigm for real-time operation.

\begin{figure}[t] 
    \centering
    \includegraphics[width=1\linewidth]{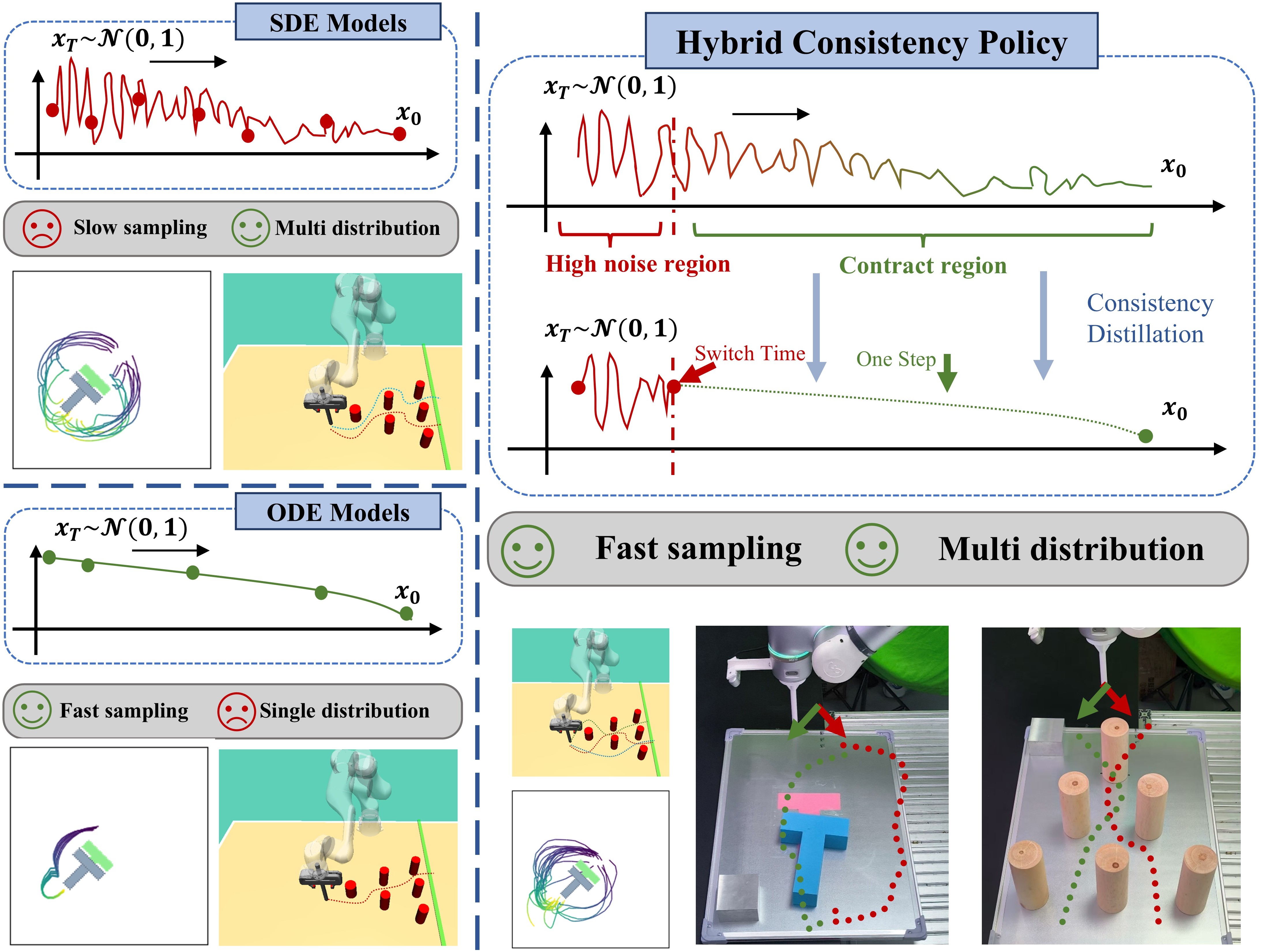}
    \caption{\textbf{Hybrid Consistency Policy.} SDE models capture multi-modal behaviors but sample slowly, while ODE models are fast yet prone to the risk of mode collapse. HCP runs a short stochastic SDE prefix in the high-noise region to form branches, then at an adaptive switch time $t_s^\ast$ performs a one-step consistency jump along the probability-flow ODE to $x_0$. This yields fast sampling and reliable multi-distribution execution, illustrated by distinct successful robot trajectories. }
    \label{fig:hcp_method}
\vspace{-5mm}
\end{figure}
A central challenge in deployment is multi-modality \cite{d3il,yu2020meta,florence2022implicit}. Real-world robotic tasks often possess highly multi-modal solution spaces due to ambiguous goals and stochastic environments \cite{li2023vision}.
As shown in Fig.~\ref{fig:hcp_method}~(left), SDE sampling injects randomness to preserve diverse behaviors, whereas deterministic ODE trajectories under distillation constraints often bias toward a dominant mode, reducing diversity and robustness~\cite{song2023improved,yin2024one,samaddar2025efficient}.
Consequently, an open problem is how diffusion-based methods can retain fast inference while preserving and promoting multi-modality in complex robotic settings.

% To address these issues, we propose the Hybrid Consistency Policy (HCP), which enables robust preservation and promotion of multiple modes during distillation. Our method introduces a principled approach for selecting distillation time steps and incorporates explicit mechanisms to encourage mode bifurcation throughout the distillation process, as shown in Fig.\ref{fig:hcp_method}. 

% We validate our approach through comprehensive experiments in both simulated environments and on real robotic platforms. The results demonstrate substantial improvements in multi-modal trajectory generation, task success rates, and sample efficiency over existing baselines.

To jointly achieve multi-modal expressiveness and real-time inference, we propose the \textbf{Hybrid Consistency Policy (HCP)}. The core idea is to run a short stochastic SDE prefix in the high-noise regime until an adaptive switching time $t_s^\ast$ at which a stable branched distribution emerges; we then perform a single consistency jump that yields the final action (Fig.~\ref{fig:hcp_method}, right). To precisely align this one-jump generation with the teacher distribution, we conduct time-varying consistency distillation along a continuous noise schedule. Concretely, we combine a consistency-trajectory term \(L_{\mathrm{CTM}}\) to enforce prediction agreement across adjacent time steps and a denoising matching term \(L_{\mathrm{DSM}}\) to improve pointwise fidelity. The switching time \(t_s^\ast\) is chosen based on changes in the teacher model’s noise distribution.

We evaluate HCP systematically on both simulated and real-world platforms. On simulation tasks \cite{d3il,chen2023diffusion}, HCP attains an average success rate of 75.5\% with a distribution entropy of 1.50, approaching the multi-modal capacity of DP. On real-world tasks, HCP reduces the per-sequence action generation time from 0.54\,s to 0.17\,s on an RTX 4080 GPU, while maintaining comparable overall success rates and multi-modal behavior to DP. These results demonstrate that HCP delivers significant real-time gains without notably sacrificing accuracy or multi-modal capability.

Our main contributions are:
\begin{itemize}
    \item We present a general framework for mode-preserving policy distillation in diffusion-based robotics, with practical strategies for time step selection and explicit mode promotion.
    \item We introduce a switch-time selection criterion based on stable mode bifurcations to stabilize accuracy and improve inference efficiency without sacrificing diversity.
    \item We extensively validate HCP on diverse simulation benchmarks and 6 real-world tasks, achieving up to a 68\% reduction in action generation time while maintaining competitive success rates and multi-modal coverage. This demonstrates faster inference and a key advantage over conventional stochastic methods.
\end{itemize}

%% file: sections/2_Relatedworks.tex
\section{Related Works} \label{Sec: related}

\subsection{Diffusion Policy in Robot Manipulation}  
Diffusion-based policies have demonstrated strong performance in learning multi-modal action distributions for robot manipulation tasks \cite{mahmoudi2024leveraging}. Diffusion Probabilistic Models (DDPMs) \cite{chi2023diffusion} work by progressively denoising random noise to generate actions that align with expert demonstrations. While DDPMs provide high-quality action generation, the iterative nature of the diffusion process results in significant computational overhead, making them impractical for real-time applications~\cite{ho2020denoising}. 

To address this limitation, Denoising Diffusion Implicit Models (DDIMs) \cite{song2020denoising} have been developed as a more efficient alternative. These models reduce the number of denoising steps by introducing a deterministic process that preserves the quality of generated actions while decreasing inference time. However, despite the efficiency improvements, both DDPMs and DDIMs still struggle with maintaining diverse behavior, particularly in environments where multiple action sequences can achieve the same task, leading to the challenge of mode collapse \cite{reuss2023goaldiffusion}.

Recent advancements, such as consistency distillation methods, aim to address these limitations by using ODEs to perform one-step generative sampling, enabling faster inference~\cite{zhai2024motion}. These models preserve the diversity of behaviors while significantly reducing computational complexity, making them suitable for real-time robotic control \cite{sun2025flashback}. 

\subsection{Consistency Distillation}

Distillation has become a key approach for accelerating ODE-based policies \cite{meng2023distillation}. Progressive distillation \cite{kim2023consistency} compresses the number of steps by distilling a slow teacher into a fast student while largely preserving sample quality. Consistency Model (CM) \cite{pmlr-v202-song23a} is distilled to enable one-step generation by enforcing agreement of denoising results that start from different locations along the same ODE trajectory. This approach has been extended to the robotics domain. Consistency Policy \cite{consistency} distills DP into a few-step student and accelerates inference by about an order of magnitude while retaining competitive success rates. FlashBack \cite{sun2025flashback} achieves one-step denoising, significantly reducing the iterations required for action generation while preserving high fidelity to the original distribution. These methods distill the generative process into a more efficient form by leveraging the inherent ODE structures of diffusion models \cite{zheng2024trajectory}.

% Consistency distillation methods have recently gained attention as a promising solution to the slow inference times of traditional diffusion-based policies \cite{meng2023distillation}. These methods distill the generative process into a more efficient form by leveraging the inherent ODE structures of diffusion models \cite{zheng2024trajectory}. CMs utilize a distilled probability flow that enables one-step denoising, dramatically reducing the number of iterations needed for generating actions while maintaining high fidelity to the original distribution~\cite{pmlr-v202-song23a}.

The key advantage of consistency distillation is its ability to speed up inference without compromising the quality of the output~\cite{peyre2019computational}. CM has shown significant improvements in both task success rates and sample efficiency, making it a valuable tool for high-performance robotic manipulation.

%% file: sections/3_Preliminary.tex
\section{Preliminary} \label{Sec: prelim}

\textbf{Problem Formulation}.  
We consider a robot manipulation task where the robot must execute a series of actions based on visual observations, to learn a policy from expert demonstrations \cite{singh2022reinforcement}. Let \( s_t \in \mathcal{S} \) be the latent state, \( o_t \in \mathcal{O} \) be the robot's observation (e.g., images and proprioception), and \( a_t \in \mathcal{A} \) be the control action (e.g., end-effector pose). The task is to learn a policy \( \pi_\theta(a_t| o_t) \) that maximizes the likelihood of expert demonstrations while accounting for ambiguities in the task. These ambiguities arise from the existence of multiple valid behaviors for the same task, creating a multi-modal distribution of actions. Thus, the goal is to train a policy that not only imitates the expert behavior but also preserves multiple modes during inference.

\textbf{Consistency Model}.  
The CM in generative modeling leverages the framework of score matching and ODEs to transform a noisy sample into a clean one. It is based on the observation that the generative process in diffusion models can be interpreted as a series of steps along an ODE trajectory, to map different noisy states at various time steps back to the same clean starting point. As the noise is gradually removed, the process follows a Probability Flow ODE (PF-ODE):
\begin{equation}
\label{eq:pf-ode}
\dot{x}_t = \tilde{f}_\theta(x_t, t),
\end{equation}
where \( \tilde{f}_\theta(x_t, t) \) is the deterministic flow function, which describes how the noisy state evolves towards the clean state.

The core idea behind the CM is to enforce self-consistency along these trajectories by training a model to predict the same clean state from different noisy points along the same ODE trajectory. More formally, given two points \( (x_{t_i}, t_i) \) and \( (x_{t_j}, t_j) \), where \( t_i > t_j \), the model is trained to ensure that the denoised predictions at both points align when mapped to a clean state. The loss function for this consistency is typically defined as:
\begin{equation}
\mathcal{L}_{\text{CM}} = \mathbb{E}\left[\| g_\phi(x_{t_i}, t_i, c) - g_\phi(x_{t_j}, t_j, c) \|^2 \right],
\end{equation}
where \( g_\phi \) represents the neural network model that predicts the clean state at different time steps, and \( c \) denotes conditioning on the observation or task.

The CM can thus generate samples in fewer steps, significantly speeding up inference times compared to the iterative denoising processes used in traditional diffusion models.

\begin{figure*}[t] 
    \centering
    \includegraphics[width=0.97\linewidth]{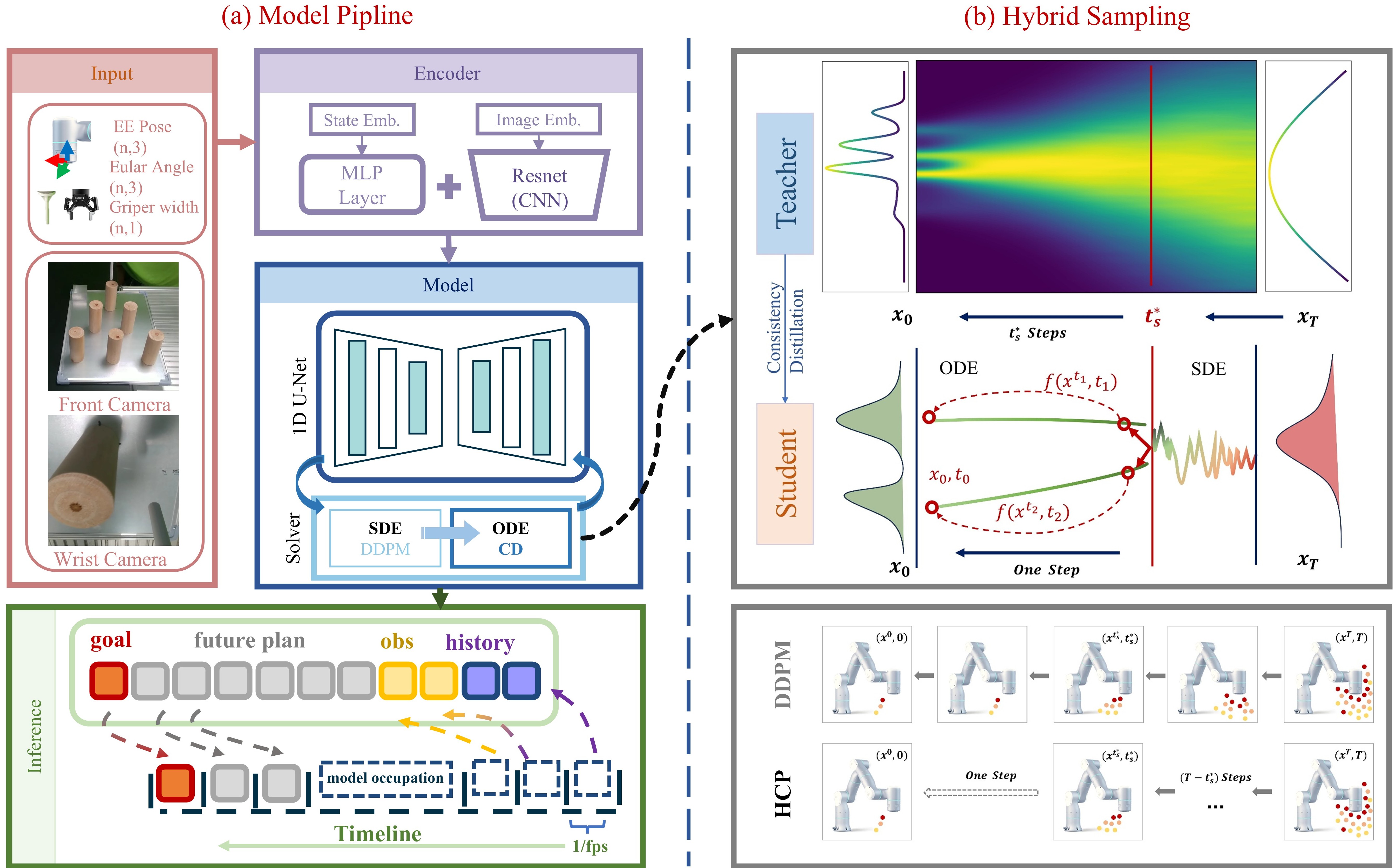}
    \caption{\textbf{Overview of the HCP architecture.} (a) Policy pipeline: robot state and multi-view images are encoded by MLP and ResNet. Inference executes action steps at equal time intervals through an action chunk. (b) Hybrid sampling: a DDPM teacher supplies stochastic trajectories, while a student is trained via consistency distillation to satisfy a one-step ODE mapping in the contract region.}
    \label{fig:hcp_overview}
\vspace{-5mm}
\end{figure*}
% \begin{figure}[t]
%     \centering
%     \includegraphics[width=0.97\columnwidth]{figures/single_column.png}
%     \caption{prelim}
%     \label{Fig: prelim}
% \end{figure}

%% file: sections/4_Method.tex
\section{Method: Hybrid Consistency Policy} \label{Sec: method}

We present HCP, which couples a stochastic prefix governed by a reverse SDE with a one-step distillation, retaining multi-modal branching while achieving fast inference. 

We first define a hybrid score matching model that parameterizes both dynamics under a shared time and noise schedule, providing a continuous bridge from the SDE regime to the ODE regime. We then carry out time-varying consistency distillation, which maps an intermediate noisy state at time $t_s$ to the clean action \cite{consistency}. Finally, we identify the switching time that marks the handoff from the stochastic prefix to the deterministic jump so that mode selection occurs before the jump. An overview of the method is shown in Fig.~\ref{fig:hcp_overview}.

\subsection{Hybrid Score Matching Model}
\label{sec:method:hsm}

We model data trajectories \(x_t\) on \(t \in [0,T]\) with a reverse-time SDE, as described in Eq.~\eqref{eq:sde}, and its PF-ODE as shown in Eq.~\eqref{eq:pf-ode}. The stochasticity of SDE supports multi-modal behaviors, while ODE is deterministic and may average modes. Hence, HCP executes a stochastic prefix followed by a one-step jump as in Eq.~\eqref{eq:pipeline}:
\begin{equation}
\label{eq:sde}
    \mathrm{d}x_t = f_\theta(x_t,t)\,\mathrm{d}t + g(t)\,\mathrm{d}w_t,
\end{equation}
% \begin{equation}
% \label{eq:pfode}
%     \dot{x}_t = \tilde f_\theta(x_t,t)，
% \end{equation}
\begin{equation}
\label{eq:pipeline}
x_{t=T}\xrightarrow[\text{$N_s$ steps}]{\text{DDPM}} x_{t_s}
\quad\Rightarrow\quad
x_0 \approx g_\phi\!\left(x_{t_s}, t_s, c\right),
\end{equation}
where \(c\) is the conditioning and \(t_s\) is selected in Sec.~\ref{sec:method:ts}.

We use a DDPM teacher trained with noise prediction and parameterize the student in a continuous \((\alpha(t), \sigma(t))\) domain. Let \(t_{\text{DDPM}} \in \{0, \dots, K\}\) be the teacher step. We define the cumulative \(\alpha_{\text{cum}}(t)\) as in Eq.~\eqref{eq:alphacum} and align discrete steps to the student’s schedule using Eqs.~\eqref{eq:align1} and \eqref{eq:align2}. Given the teacher’s noise estimate \(\hat\epsilon(x_t,t)\), the denoised target is given by Eq.~\eqref{eq:eps-to-x0}:
\begin{equation}
\label{eq:alphacum}
\alpha_{\text{cum}}(t) = \prod_{i=0}^{t} \alpha_i = \prod_{i=0}^{t}(1-\beta_i),
\end{equation}
\begin{equation}
\label{eq:align1}
\alpha(t_k) = \sqrt{\alpha_{\text{cum}}(t_k)}, 
\end{equation}
\begin{equation}
\label{eq:align2}
\sigma(t_k) = \sqrt{\frac{1-\alpha_{\text{cum}}(t_k)}{\alpha_{\text{cum}}(t_k)+\varepsilon}},
\end{equation}
\begin{equation}
\label{eq:eps-to-x0}
x_{t-1} = x_t - \sigma(t) \hat\epsilon(x_t,t).
\end{equation}
where \(\varepsilon > 0\) is a small constant for numerical stability. 

In practice, we compute \(\alpha_{\text{cum}}\) from the DDPM scheduler, map it to \((\alpha, \sigma)\), run \(t_s\) reverse-DDPM steps to obtain \(x_{t_s}\), and apply one evaluation of \(g_\phi(x_{t_s}, t_s, c)\) to predict \(x_0\).

\subsection{Time-varying Consistency Distillation}
\label{sec:method:distill}

We operate in the continuous noise-parameterized schedule (e.g., \(\sigma\)-domain) and sample a triplet of times \((s,u,t)\) from the scheduler as in Eq.~\eqref{eq:sampling_triple}. Given a noisy state \(x_t\) at time \(t\), we obtain its predecessors \(x_u\) and \(x_s\) along the reverse trajectory (e.g., by teacher stepping or cached prefixes). The teacher predicts noise \(\hat\epsilon(x_\tau,\tau)\) for any \(\tau \in \{s,u,t\}\), using the standard \(\epsilon \to x_0\) conversion as in Eq.~\eqref{eq:eps-to-x0}. The student \(g_\phi\) directly outputs an \(x_0\) estimate from \((x_\tau,\tau,c)\). For brevity, we denote:
\begin{equation}
\label{eq:sampling_triple}
\sigma_{\min} \le s \le u \le t \le \sigma_{\max}.
\end{equation}
\begin{equation}
\label{eq:epsfg}
f(x_\tau,\tau) \triangleq g_\phi(x_\tau,\tau,c).
\end{equation}

To ensure that the student's predictions are temporally coherent across nearby steps, we minimize the discrepancy between its outputs at \(s\) and \(u\), as shown in Eq.~\eqref{eq:l_ctm_new}:
\begin{equation}
\label{eq:l_ctm_new}
L_{\text{CTM}} = \left\| f(x_s,s) - f(x_u,u) \right\|_2^2.
\end{equation}
where both \(f(x_s,s)\) and \(f(x_u,u)\) should be close to the same underlying clean target that the teacher would reveal at \(t\). 
% Enforcing their agreement stabilizes the local temporal neighborhood and reduces sensitivity to the precise switching time.

The DSM term evaluates how well the student reconstructs the clean sample, as shown in Eq.~\eqref{eq:l_dsm_new}:
\begin{equation}
\label{eq:l_dsm_new}
L_{\text{DSM}} = \mathbb{E}_{t, x_0, x_t | x_0}\left[d (\hat x_0(x_t,t), x_0)\right],
\end{equation}
where \(\hat x_0(x_t,t)\) is computed via Eq.~\eqref{eq:eps-to-x0} from the teacher’s noise prediction, and \(d(\cdot, \cdot)\) is an error metric between the denoised prediction and the ground-truth trajectory (e.g., squared \(\ell_2\)).

The student is trained with a weighted combination of the two terms, as shown in Eq.~\eqref{eq:l_cd}:
\begin{equation}
\label{eq:l_cd}
L_{\text{CD}} = \alpha L_{\text{CTM}} + \beta L_{\text{DSM}},
\end{equation}
where (i) \(L_{\text{CTM}}\) enforces the CM across adjacent times, guiding the student to yield time-invariant predictions around the switch region, and (ii) \(L_{\text{DSM}}\) is the denoising score matching loss that measures the absolute denoising quality against the teacher/ground-truth signal. In practice, \(\alpha\) and \(\beta\) can be tuned to balance temporal smoothness and pointwise accuracy.

\begin{figure}[t]
    \centering
    \includegraphics[width=0.95\linewidth]{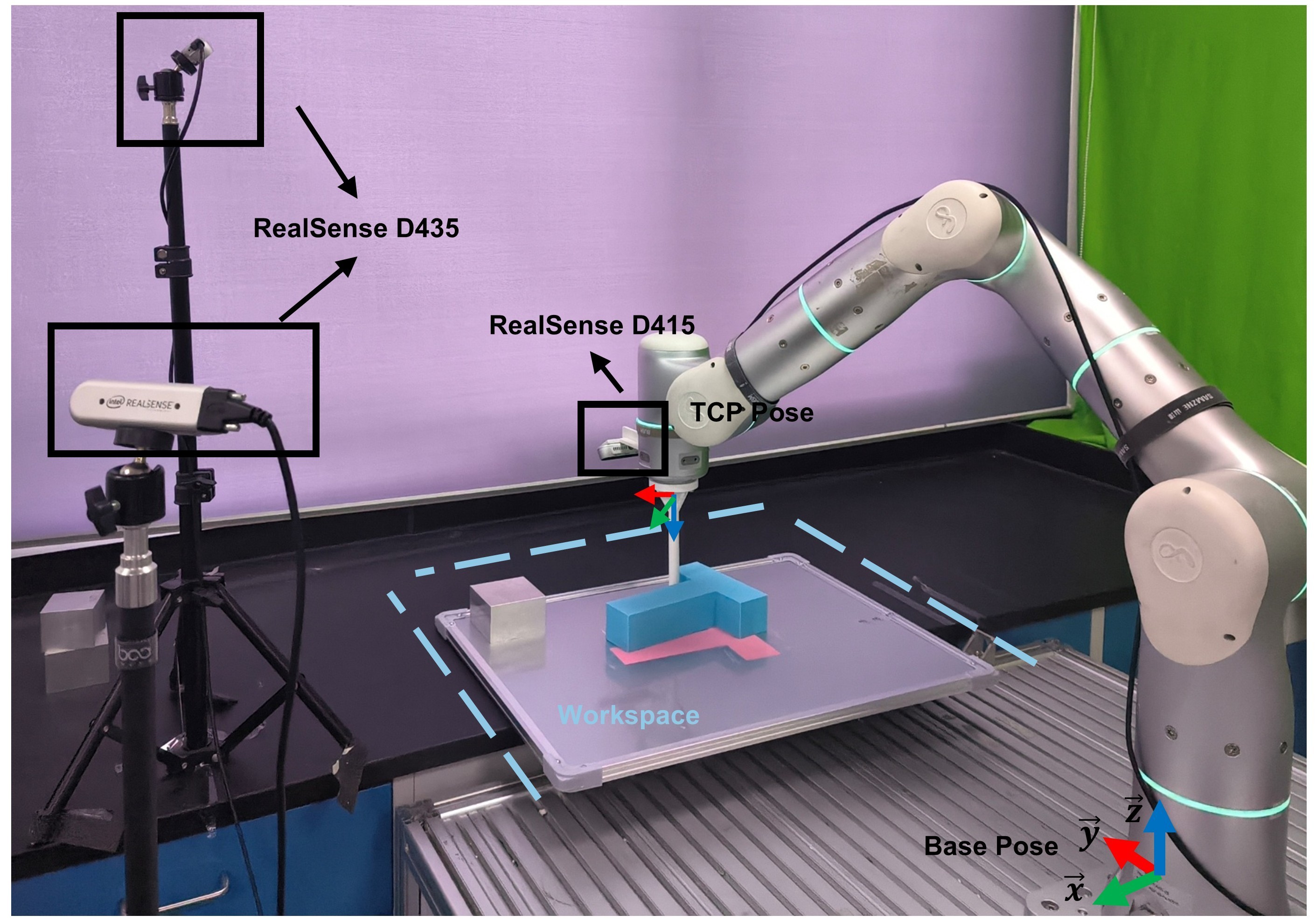}
    \caption{\textbf{Real-world setup and sensors.} 
    A 7-DoF collaborative arm operates in a fixed workspace with calibrated TCP pose. 
    A wrist camera (RealSense D415) and a third-person camera (RealSense D435) provide multi-view observations.}
    \label{fig:real_setup}
\vspace{-4mm}
\end{figure}

\begin{figure}[t]
    \centering
    \includegraphics[width=0.95\columnwidth]{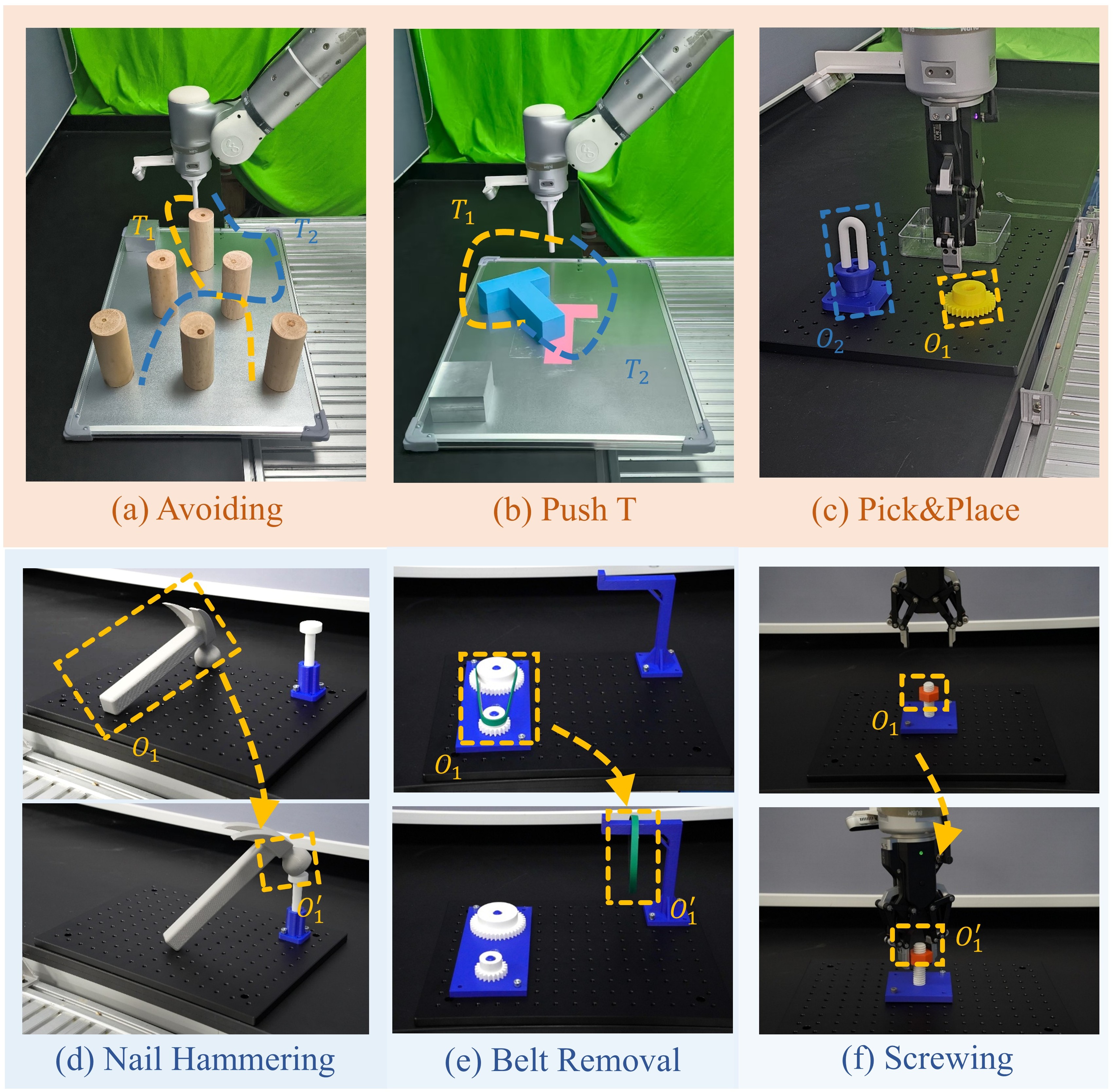}
    \caption{\textbf{Real-world tasks.} (a), (b) and (c) are multi-modal tasks; (d), (e) and (f) are single-modal tasks. All demonstrations are collected via VR teleoperation with 60 successful demos per task.}
    \label{fig:real_tasks}
\vspace{-4mm}
\end{figure}

% \subsection{Switching Time Optimization}
% \label{sec:method:ts}

% At each reverse-time step $t$, we draw $N$ samples and fit a Gaussian mixture with up to $k_{\max}$ components. The number of modes $K(t)$ is chosen by the Bayesian information criterion. After ordering the component means, we take the smallest distance between neighboring means as the inter-mode gap at step $t$. To reduce noise, we apply a short moving average over $w$ steps to obtain a smoothed gap,
% \begin{equation}
% \label{eq:smooth}
% \mathrm{gap}_s(t)=\frac{1}{w}\sum_{i=0}^{w-1}\mathrm{gap}(t+i),
% \end{equation}
% which we use to detect stable splits.

% We also compute the noise cumulative share $(1-\alpha_{\mathrm{cum}}(t))$ from the scheduler. 
% To capture distributional change between consecutive reverse steps, we use the $L_1$ difference of column-normalized KDEs:
% \begin{equation}
% \hat p_t(x)=\frac{\mathrm{KDE}_t(x)}{\max\nolimits_x \mathrm{KDE}_t(x)} ,
% \end{equation}
% \begin{equation}
% c(t)=\int \big|\hat p_t(x)-\hat p_{t-1}(x)\big|\,dx,\quad c(0)=0 .
% \end{equation}

% In practice, we fit Gaussian-kernel $\mathrm{KDE}_t$ with Scott’s rule and approximate the integral on the evaluation grid.

% We scan forward and pick the earliest $t$ that satisfies
% \begin{align}
% \label{eq:rule}
% \text{(i)}\;& \mathrm{gap}(t_s)\ge \tau\ ,\\
% \text{(ii)}\;& 1-\alpha_{\mathrm{cum}}(t_s)\le \eta \ , \nonumber\\
% \text{(iii)}\;& c(t_s)\ge (1-\gamma)\max c \ . \nonumber
% \end{align}
% The earliest such $t$ is $t_s^\ast$. Hyperparameters are $\tau,\eta,\gamma$.

\subsection{Switching Time Optimization}
\label{sec:method:ts}

At each reverse-time step \( t \), we scan forward and select the earliest \( t \) that satisfies the following three criteria:
\begin{align}
\text{(i)}\;& \mathrm{gap}(t_s) \ge \tau\ ,\\
\text{(ii)}\;& 1 - \alpha_{\mathrm{cum}}(t_s) \le \eta \ , \nonumber\\
\text{(iii)}\;& c(t_s) \ge (1 - \gamma) \max c \ . \nonumber
\end{align}

The earliest such \( t \) is denoted as \( t_s^\ast \). Hyperparameters are \( \tau \), \( \eta \), and \( \gamma \). We now explain each of these criteria.

To detect stable splits, we first draw \( N \) samples and fit a Gaussian mixture with up to \( k_{\max} \) components. The number of modes \( K(t) \) is chosen using the Bayesian information criterion. After ordering the component means, we compute the smallest distance between neighboring means, which is referred to as the inter-mode gap. To reduce noise, we apply a short moving average over \( w \) steps, as shown in Eq.~\eqref{eq:gap}:
   \begin{equation}
   \label{eq:gap}
   \mathrm{gap}_s(t) = \frac{1}{w}\sum_{i=0}^{w-1}\mathrm{gap}(t+i),
   \end{equation}

To track the noise reduction over time, we compute the cumulative share of noise \( 1 - \alpha_{\mathrm{cum}}(t) \) from the scheduler. This value indicates how much noise remains at each reverse-time step. A lower cumulative noise share suggests that the system is close to a clean state, making it a suitable time to switch.

To capture the distributional change between consecutive reverse steps, we use the \( L_1 \) difference of column-normalized Kernel Density Estimates (KDEs) at steps \( t \) and \( t-1 \), as shown in Eq.~\eqref{eq:ct_pt}:
   \begin{equation}
   \hat p_t(x) = \frac{\mathrm{KDE}_t(x)}{\max_x \mathrm{KDE}_t(x)} ,
   \end{equation}
   \begin{equation}
   \label{eq:ct_pt}
   c(t) = \int \big|\hat p_t(x) - \hat p_{t-1}(x)\big|\,dx, \quad c(0) = 0 .
   \end{equation}
   
This change is used to identify when the distribution has sufficiently adapted. In practice, we fit the Gaussian-kernel \( \mathrm{KDE}_t \) with Scott’s rule and approximate the integral on the evaluation grid.

\begin{figure*}[t]
    \centering
    \includegraphics[width=1\textwidth]{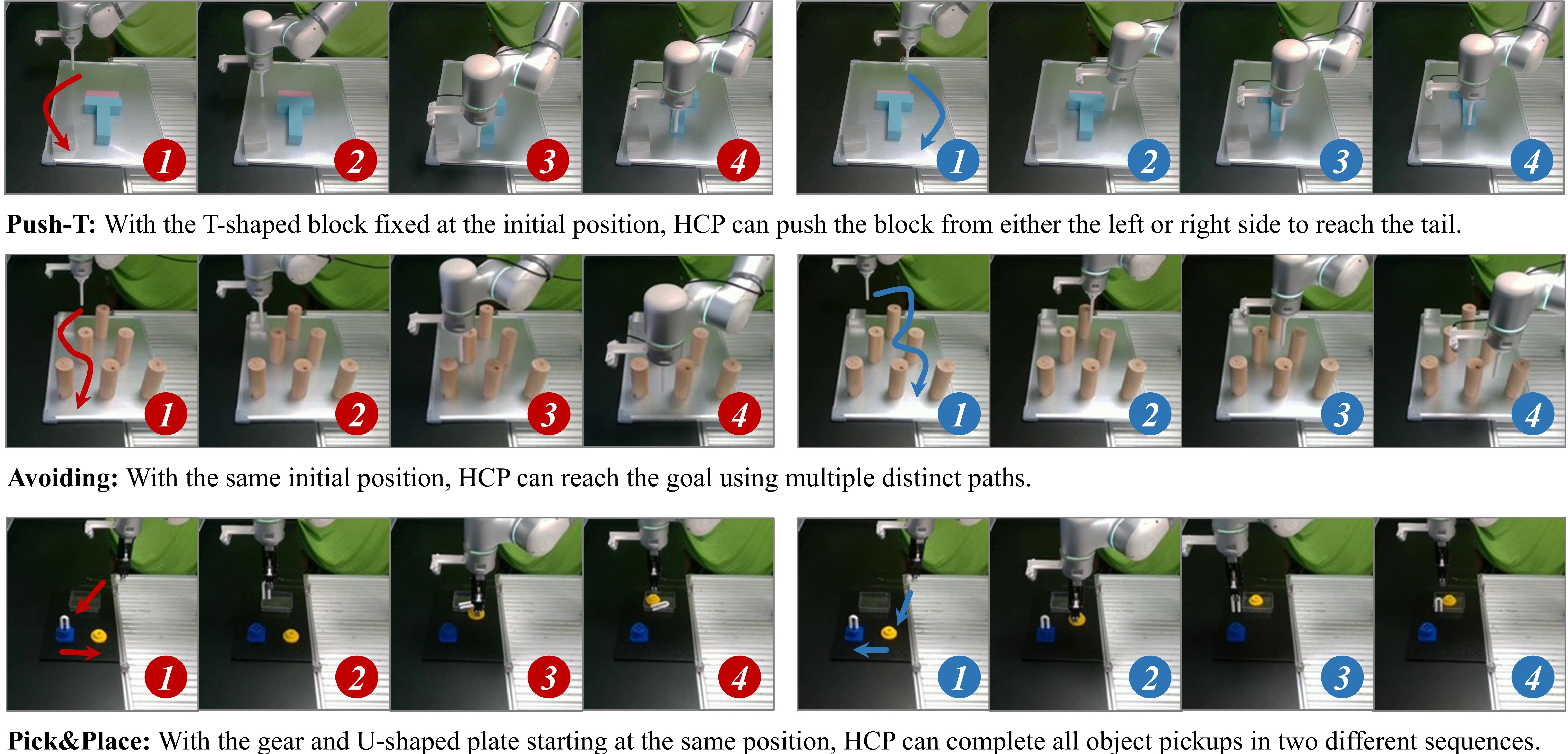}
    \caption{\textbf{Performance of HCP in real-world multi-modal tasks.} HCP successfully achieves multiple modes for each task, demonstrating performance close to the Teacher in terms of accuracy and multi-modal capabilities.}
    \label{fig:experiment1}
\vspace{-5mm}
\end{figure*}

%% file: sections/5_Experiment.tex
\section{Experiment} \label{Sec: experiment}

We evaluate the performance of our proposed HCP model in simulation on several multi-modal tasks as well as through real-robot experiments. Our experiments are designed to answer three questions:
\begin{itemize}
    \item \textbf{Q1 (Effectiveness)}: Does HCP preserve the multi-modal behavior of SDE-based diffusion policies while matching their success?
    \item \textbf{Q2 (Efficiency)}: How does the HCP improve the real-time performance compared with baselines? 
    \item \textbf{Q3 (Ablation)}: How important is reaching the switch time $t_s^\ast$? What happens if we jump early (HCP w/o $t_s^\ast$)?
\end{itemize}

% Q1: can HCP achieve multi-modal policy distribution? to DP, FM, HP
%  also visualization of results. tables.
% Q2: can switch time opt help the multi-modal?
%  a kind of ablation
% Q3: how does the HCP improve the real-time performance compared with baselines? 
% computation time comparison

% real robot

\subsection{Experimental Setup}

\textbf{Environments}.
We use a 7-DoF collaborative arm, Flexiv Rizon,  in a fixed tabletop workspace with calibrated base and tool-center point (TCP) frames (Fig.~\ref{fig:real_setup}). A wrist-mounted Intel RealSense D415 and a third-person Intel RealSense D435 provide multi-view observations. During teleoperation, the control loop runs at 30\,Hz and demonstrations are logged at 10\,Hz; during policy rollouts, the controller executes at 10\,Hz. Each real-world task contains 60 successful demonstrations collected via VR teleoperation. Real-world deployment uses the LeRobot framework \cite{cadene2024lerobot}. Inference runs on an RTX~4080 GPU with a host machine of 32\,GB RAM. All input images are cropped to 224 pixels 224 before being fed to the model. 

\textbf{Tasks}. We evaluate two multi-modal simulation tasks and six real-robot tasks (Fig.~\ref{fig:real_tasks}); all real-robot datasets contain \textbf{60} successful demonstrations collected via VR teleoperation.

\begin{itemize}
  \item \emph{Multi-modal (simulation).} \textbf{Push-T-Sim} \cite{chen2023diffusion} is a manipulation task where the arm uses the end-effector to push a T-shaped block from randomized initial poses to a fixed goal, and training uses 200 demonstrations; \textbf{Avoiding-Sim} \cite{d3il} is a motion-planning task where the arm moves from a fixed start to a green finish line while avoiding six cylinders, evaluated with the official dataset of 96 demonstrations.
  
  \item \emph{Multi-modal (real).} \textbf{Push-T-Real} is a manipulation task where the arm starts from a slightly perturbed pose and pushes a randomly posed T-block to the target region; \textbf{Avoiding-Real} is a motion-planning task from a fixed start to a finish while avoiding cylinders; \textbf{Pick\&Place} is a manipulation task where a yellow gear and a white U-shaped plate are randomly placed and grasped in random order.
  
  \item \emph{Single-modal (real).} \textbf{Nail Hammering} is a tool-usage task where the arm grasps a hammer and drives a nail whose hammer and nail poses are randomized; \textbf{Belt Removal} is an assembly task where the arm grasps the belt on a gear plate and mounts it onto a fixed bracket; \textbf{Screwing} is a precision task where the arm grasps a bolt and rotates it by about \textbf{60\%}.
\end{itemize}

\textbf{Baselines}.
We compare two standard diffusion policies and two HCP variants under the same backbone and training pipeline: DP-ddpm (80 steps), DP-ddim (50 steps), HCP (ours) with a 25-step SDE prefix ($>t_s^\ast$), and HCP w/o $t_s^\ast$ with a 5-step SDE prefix. All methods take two consecutive observations as input and output an eight-step end-effector pose sequence.

\textbf{Metrics}. We evaluate three aspects. (i) Effectiveness measures episode success rate \textbf{Acc.} in percent for each task. (ii) Efficiency uses the number of denoising function evaluations (\textbf{NFE}) and wall clock time in seconds on a single RTX~4080, reporting \textbf{Time} to generate one action sequence for multi-modal tasks and \textbf{Total Time} per episode for the single-modal tasks. (iii) Multi-modality uses the number of solution families (\textbf{Modes}) and the Shannon \textbf{Entropy} \(H(\mathbf{p}) = -\sum_{i=1}^{n} p_i \log_2 p_i\) where \(p_i\) is the empirical frequency of family \(i\).

We run 100 trials per simulation task and 20 trials per real robot task. Higher is better for Acc., Entropy, and Modes. Lower is better for NFE and Time.

\begin{figure}[t]
    \centering
    \includegraphics[width=1\columnwidth]{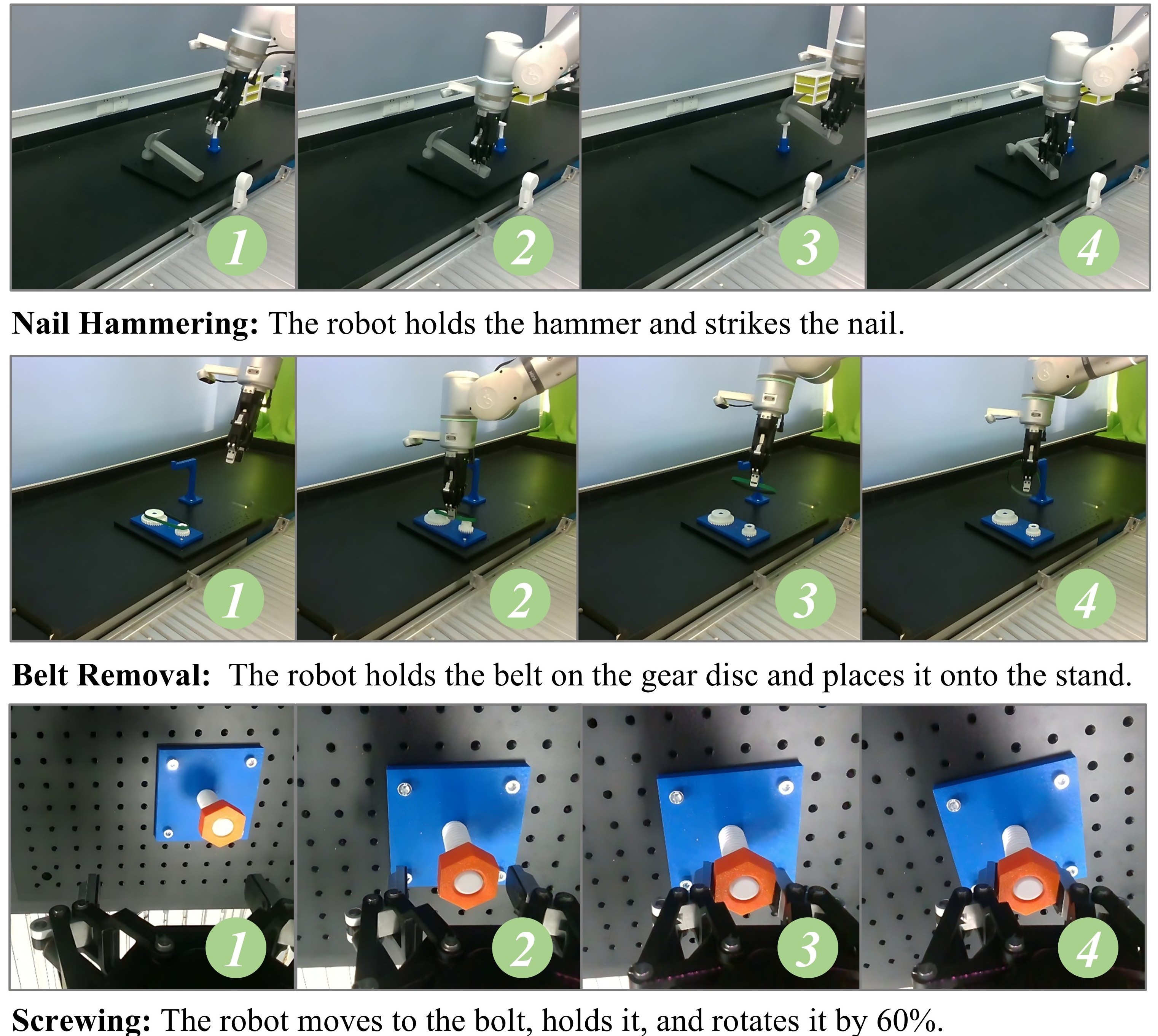}
    \caption{\textbf{Performance of HCP in real-world single-modal tasks.} HCP performs comparably to the Teacher in tasks like tool manipulation and object grasping, maintaining high accuracy.}
    \label{fig:experiment2}
\vspace{-5mm}
\end{figure}

\begin{table*}[t]
    \centering
    \caption{\textbf{Main Results In Simulated Multi-Modal Tasks.}}
    \label{tab:main_sim}
    % Make the table span single column width
    \small
    \setlength{\tabcolsep}{7pt}  % Reduce column spacing
    \renewcommand{\arraystretch}{1.15}  % Adjust row height
    \begin{tabular}{l c c c c c c c c c c c}
        \toprule[1pt]
        \multirow{2}{*}{Methods} & \multirow{2}{*}{NFE$\downarrow$} & \multicolumn{3}{c}{Push-T-Sim} & & \multicolumn{3}{c}{Avoiding-Sim} & & \multicolumn{2}{c}{Overall} \\
        \cline{3-5}
        \cline{7-9}
        \cline{11-12}
        & & Acc.$\uparrow$ & Entropy$\uparrow$ & Mode$\uparrow$ & & Acc.$\uparrow$ & Entropy$\uparrow$ & Mode$\uparrow$ & & Acc.$\uparrow$ & Entropy$\uparrow$ \\
        \hline
        DP-ddpm & 80 & 69\% & 0.99 & 2 & & 89\% & 2.52 & 6 & & 79\% & 1.76 \\
        DP-ddim & 50 & 74\% & 0 & 1 & & 100\% & 0 & 1 & & 87\% & 0 \\
        HCP w/o $t_s^\ast$ & 5 & 59\% & 0.47 & 2 & & 86\% & 0 & 1 & & 72.5\% & 0.23 \\
        HCP (ours) & 25 & 64\% & 0.92 & 2 & & 87\% & 2.08 & 5 & & 75.5\% & 1.50 \\
        \bottomrule[1pt]
    \end{tabular}
\end{table*}

\begin{table*}[t]
    \centering
    \caption{\textbf{Main Results in Real-World Multi-Modal Tasks.}}
    \label{tab:main_real1}
    % Use a smaller font size to ensure the table fits within the column
    \small
    \setlength{\tabcolsep}{4.0pt}  % Reduce column spacing
    \renewcommand{\arraystretch}{1.15}  % Adjust row height
    \begin{tabular}{l c c c c c c c c c c c c c c c}
        \toprule[1pt]
        \multirow{2}{*}{Methods} &  \multirow{2}{*}{Time(s)$\downarrow$} & \multicolumn{3}{c}{Push-T-Real} & & \multicolumn{3}{c}{Avoiding-Real} & & \multicolumn{3}{c}{Pick\&Place} & & \multicolumn{2}{c}{Overall} \\
        \cline{3-5}
        \cline{7-9}
        \cline{11-13}
        \cline{15-16}
        & & Acc.$\uparrow$ & Entropy$\uparrow$ & Mode$\uparrow$ & & Acc.$\uparrow$ & Entropy$\uparrow$ & Mode$\uparrow$ & & Acc.$\uparrow$ & Entropy$\uparrow$ & Mode$\uparrow$ & & Acc.$\uparrow$ & Entropy$\uparrow$ \\
        \hline
        DP-ddpm & 0.54 & 70\% & 0.94 & 2 & & 90\% & 1.69 & 4 & & 75\% & 0.84 & 2 & & 78.33\% & 1.49 \\
        DP-ddim & 0.34 & 70\% & 0.75 & 2 & & 100\% & 0 & 1 & & 60\% & 0.41 & 2 & & 76.67\% & 0.39 \\
        HCP w/o $t_s^\ast$ & 0.04 & 5\% & 0 & 1 & & 70\% & 1.28 & 3 & & 40\% & 0.81 & 2 & & 38.33\% & 0.36 \\
        HCP (ours) & 0.17 & 55\% & 0.94 & 2 & & 90\% & 1.73 & 4 & & 75\% & 0.84 & 2 & & 73.33\% & 1.17 \\
        \bottomrule[1pt]
    \end{tabular}
\end{table*}

\begin{table*}[t]
    \centering
    \caption{\textbf{Main Results in Real-World Single-Modal Tasks.}}
    \label{tab:main_real2}
    % Use a smaller font size to ensure the table fits within the column
    \small
    \setlength{\tabcolsep}{4.9pt}  % Reduce column spacing
    \renewcommand{\arraystretch}{1.15}  % Adjust row height
    \begin{tabular}{l c c c c c c c c c c c}
        \toprule[1pt]
        \multirow{2}{*}{Methods} & \multicolumn{2}{c}{Nail Hammering}& & \multicolumn{2}{c}{Belt Removal} & & \multicolumn{2}{c}{Screwing} & & \multicolumn{2}{c}{Overall} \\
        \cline{2-3}
        \cline{5-6}
        \cline{8-9}
        \cline{11-12}
        & Acc.$\uparrow$ & Total Time (s)$\downarrow$ & & Acc.$\uparrow$ & Total Time (s)$\downarrow$ & & Acc.$\uparrow$ & Total Time (s)$\downarrow$ & & Acc.$\uparrow$ & Total Time (s)$\downarrow$\\
        \hline
        DP-ddpm (teacher)  & 50\% & 22.4 & & 80\% & 19.6 & & 90\% & 12.6 & & 73.33\% & 18.57 \\
        HCP (student) & 35\% & 19.8 & & 80\% & 14.4 & & 95\% & 9.1 & & 70\% & 14.43 \\
        \bottomrule[1pt]
    \end{tabular}
\end{table*}

\begin{figure}[t]
  \centering
  \includegraphics[width=0.97\columnwidth]{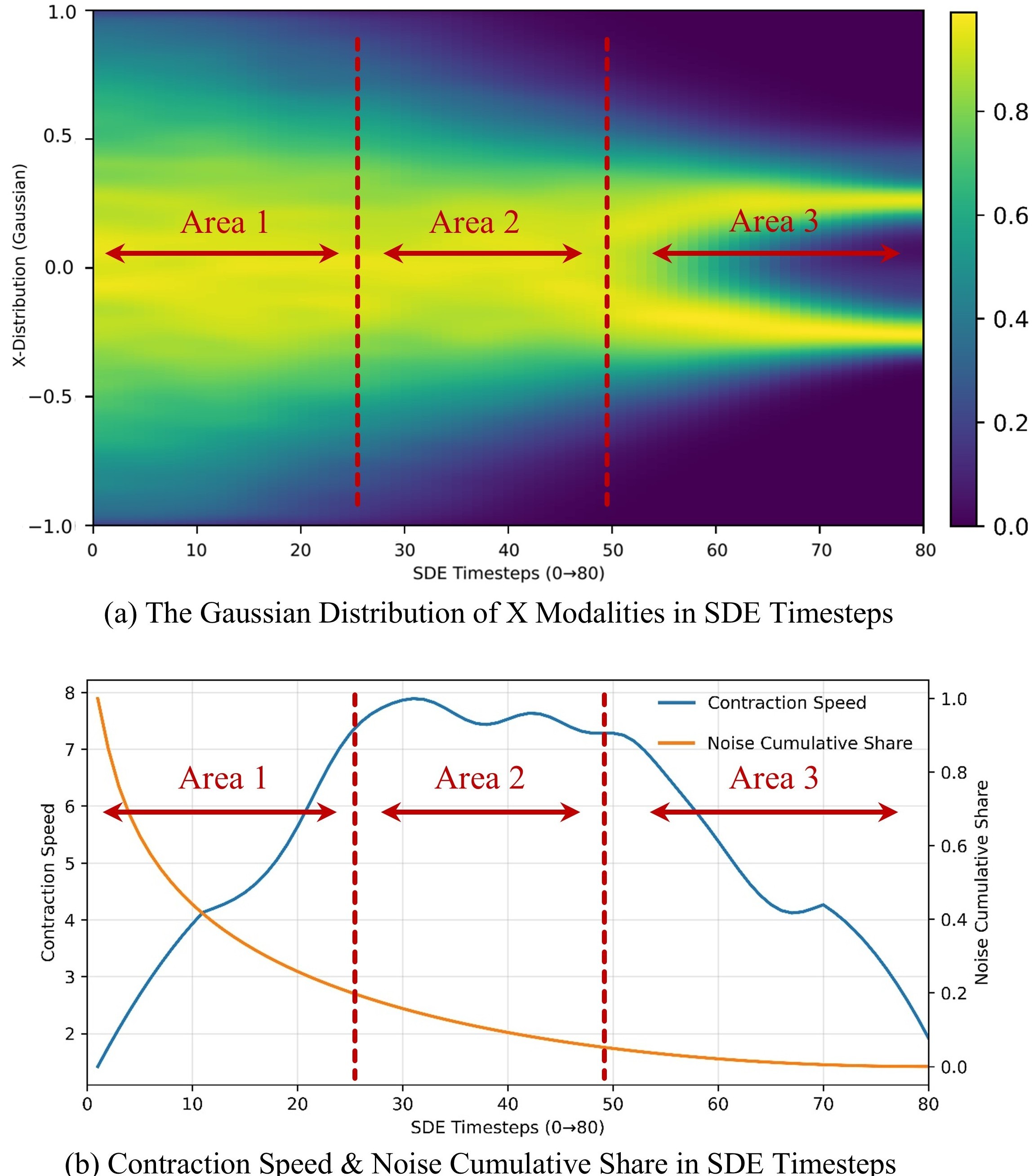}
  \caption{\textbf{SDE prefix phases and the viable switch window}. 
  (a) Evolution of modality formation over SDE steps, dashed lines split three phases: pre-bifurcation (Area 1), stabilized bifurcation (Area 2), and late contraction (Area 3). 
  (b) Contraction speed and noise cumulative share over the same steps, the peak speed together with low noise indicates the feasible window for \(t_s^\ast\) around steps 15 to 25.}
  \label{fig:areas}
\vspace{-5mm}
\end{figure}

\subsection{Simulation Results}

We evaluate the performance of the HCP in simulation, comparing it to two established baseline methods: DP-ddpm and DP-ddim. Our primary objectives are to assess the effectiveness and efficiency of HCP in preserving multi-modal behaviors and improving real-time performance.

HCP demonstrates a strong ability to preserve the multi-modal behaviors characteristic of SDE-based diffusion policies, while maintaining competitive task success rates (\textbf{Answering Q2}). As shown in Table.~\ref{tab:main_sim}, HCP performs comparably to DP-ddpm in terms of accuracy, and outperforms DP-ddim both in accuracy and multi-modal coverage. Specifically, in the Push-T-Sim task, HCP achieves a slightly lower success rate compared to DP-ddpm, but excels in multi-modal coverage. In contrast, Avoiding-Sim sees HCP achieve 87\% success, with an entropy score of 2.08 and coverage of 5 modes, which mirrors DP-ddpm’s multi-modal capacity while achieving nearly identical task success (89\%).

In terms of real-time performance, HCP outperforms baseline methods by significantly reducing inference time without compromising task accuracy (\textbf{Answering Q2}). As demonstrated in Table.~\ref{tab:main_sim}, HCP achieves a dramatic reduction in NFE. While DP-ddpm requires 80 steps and DP-ddim uses 50 steps per prediction, HCP reduces this to just 25 steps. For example, in Push-T-Sim, HCP reduces NFE from 80 to 25, while maintaining a similar multi-modal distribution (0.92 vs. 0.99) and a competitive success rate (64\% vs. 69\%).

\subsection{Real Robot Results}

To further validate HCP's performance, we test it on a real robotic system using a set of multi-modal and conventional tasks. These experiments assess both HCP’s ability to retain multi-modal capabilities and its accuracy in real-world settings. The results are presented in Fig.~\ref{fig:experiment1} and Fig.~\ref{fig:experiment2}.

HCP successfully preserves multi-modal behaviors in real-world tasks, demonstrating similar effectiveness to our simulation results (\textbf{Answering Q1}). As shown in Table.~\ref{tab:main_real1}, HCP closely matches DP-ddpm’s performance in multi-modal tasks, surpassing DP-ddim in both task success rate and mode coverage. For conventional tasks, such as those outlined in Table.~\ref{tab:main_real2}, HCP performs nearly on par with DP-ddpm in terms of success rate, but significantly accelerates task execution time. Despite this, for dynamic object manipulation tasks like Push-T-Real and Nail Hammering, HCP still lags behind DP-ddpm, highlighting areas for further improvement.

HCP excels in reducing task completion times without sacrificing accuracy, proving to be a highly efficient solution in real-world robotic applications (\textbf{Answering Q2}). As demonstrated in Table.~\ref{tab:main_real1}, HCP’s inference time is substantially lower than DP-ddpm’s, with Push-T-Real taking just 0.17 seconds for HCP, compared to 0.54 seconds for DP-ddpm. Additionally, in the Belt Removal task, HCP reduces the overall task time by 26\%. This acceleration, combined with maintained accuracy and multi-modal capabilities, establishes HCP as an ideal candidate for deployment in real-world robotic systems where speed is crucial.

\subsection{Ablation Studies}
\begin{figure}[t]
    \centering
    \includegraphics[width=0.95\columnwidth]{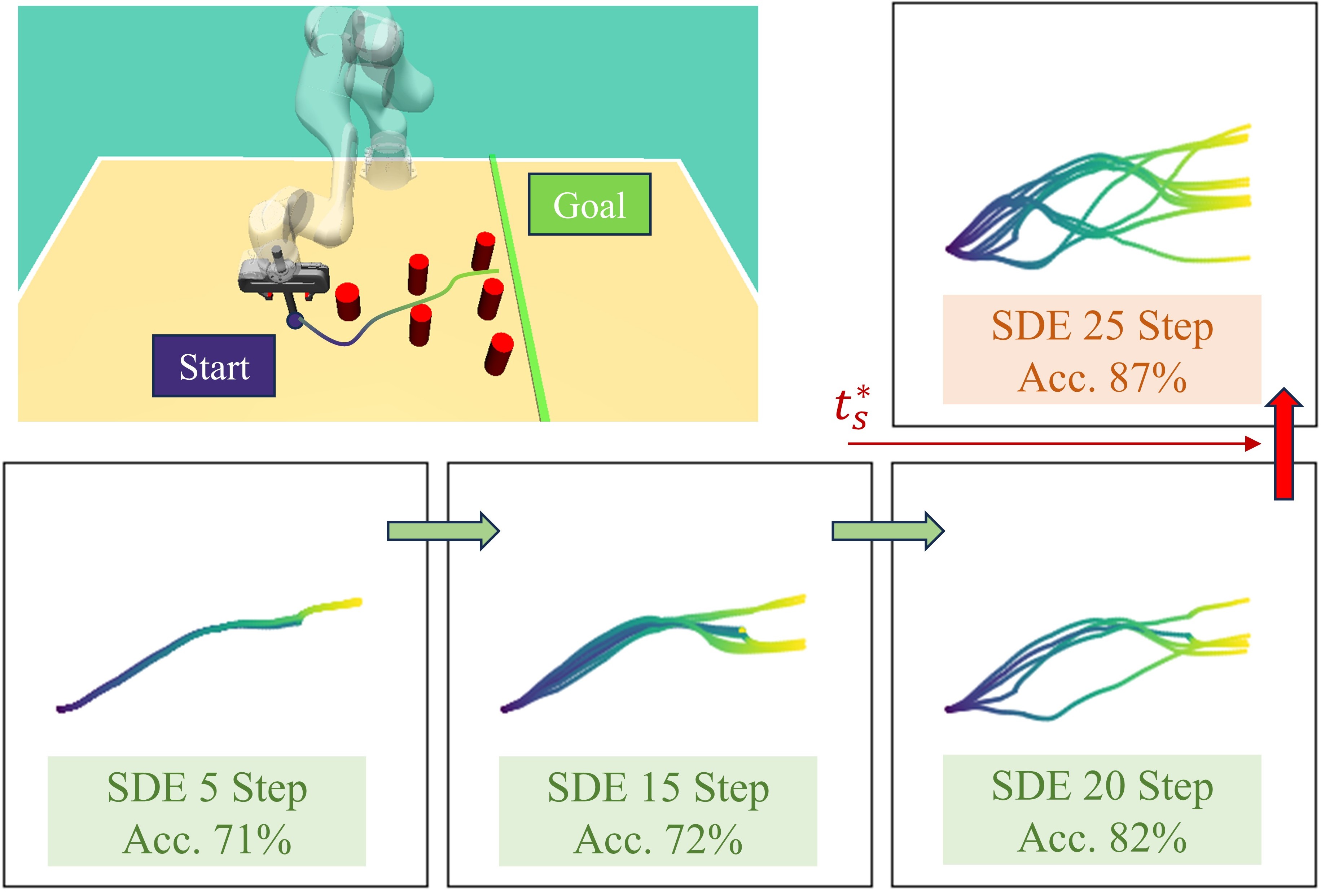}
    \caption{\textbf{Multi-modal and optimal switch time in Avoiding-Sim.} After $t_s^\ast$, success is highest and branches are stable; we then apply a one-step consistency jump to obtain final actions.}
    \label{fig:ablation_ts}
\vspace{-5mm}
\end{figure}

Early switching reduces success and collapses modes. 
Choosing \(t_s\) at or after the critical time \(t_s^\ast\) preserves diversity and feasibility (\textbf{Answering Q3}). We ablate the switch time by varying only the SDE prefix length and comparing two HCP variants: an early switch HCP w/o \(t_s^\ast\) with a 5-step prefix, and HCP (ours) with a 25-step prefix chosen beyond the feasible window of \(t_s^\ast\) (Fig.~\ref{fig:areas}, \ref{fig:ablation_ts}). 

In Push-T-Sim, HCP raises accuracy from 59\% to 64\%, increases entropy from 0.47 to 0.92, and keeps two modes while NFE grows from 5 to 25. In Avoiding-Sim, the improvement is stronger accuracy from 86\% to 87\%, entropy from 0 to 2.08. It indicates that switching too early collapses branches before bifurcation stabilizes. The pattern transfers to the real world. HCP preserves multi-modality with modest latency; in Push-T-Real, accuracy jumps from 5\% to 55\% (entropy 0 to 0.94), in Avoiding-Real it sustains 90\% with higher entropy, and in Pick\&Place accuracy improves from 40\% to 75\% (entropy 0.81 to 0.84). Overall accuracy climbs from 38.33\% to 73.33\% while per–action time rises from 0.04\,s to 0.17\,s, showing that switching after \(t_s^\ast\) trades slight latency for large gains in success and diversity.

As shown in Fig.~\ref{fig:areas}, the SDE process has three phases. Area~1 is pre-bifurcation and dominated by noise. Area~2 is a stabilized bifurcation phase. Contraction speed reaches its peak, and noise has decayed. Area~3 is a late contraction phase. Fig.~\ref{fig:ablation_ts} visualizes rollouts in the avoiding task with \(t_s\) from 5 to 25. Switching in Area~1 is premature. HCP has not formed clear modal branches. A forced jump can collapse modes. We expect an optimal switch time \(t_s^\ast\). The current estimate uses hand-tuned hyperparameters and lacks a full theory. Empirically, with an 80-step teacher, a 25-step prefix yields stable modal bifurcation. It lowers computation and latency while keeping success and coverage nearly unchanged.

%% file: sections/6_Conclusion.tex
\section{Conclusion} \label{Sec: conclusion}

By tackling the long-standing challenge of balancing multimodality and efficiency, this work provides both methodological innovation and practical insights. We presented HCP, a hybrid policy that runs a stochastic SDE prefix and then performs a one-step consistency jump along the probability flow ODE. This design retains multi-modal branching where it matters and then converges quickly, effectively decoupling mode preservation from inference speed. On Push-T and Avoiding, in simulation and on a real robot, the method approaches the DDPM teacher in success and mode coverage while cutting latency by large margins. Beyond these results, HCP provides a general recipe for accelerating diffusion-based policies while preserving their expressive capacity, enabling applications in embodied AI assistants and laying the foundation for scalable, low-latency generative control.